\documentclass[final,1p,times,twocolumn]{elsarticle}

\usepackage{amssymb}
\usepackage{amsmath}

\usepackage{comment}

\journal{Pattern Recognition}

\begin{document}

\begin{frontmatter}



\title{Bringing RGB and IR Together: Hierarchical Multi-Modal Enhancement for Robust Transmission Line Detection}


\author[1]{Shengdong Zhang}
\affiliation[1]{organization={Shaoxing University},
	addressline={Yuecheng},
	city={Shaoxing},
	postcode={312000},
	state={Zhejiang},
	country={China}}
\ead{shengdong1986@gmail.com}






\author[2]{Xiaoqin Zhang}
\affiliation[2]{organization={Wenzhou University},
	addressline={Education Park Zone},
	city={Wenzhou},
	postcode={325035},
	state={Zhejiang},
	country={China}}


\ead{zhangxiaoqinnan@gmail.com}


\author[4]{Wenqi Ren}

\affiliation[4]{organization={Sun Yat-sen University},
	addressline={Guangming district},
	city={Shenzhen},
	postcode={518000},
	state={Guangdong},
	country={China}}

\ead{rwq.renwenqi@gmail.com}

\author[5]{Linlin Shen}

\affiliation[5]{organization={Shenzhen University},
	addressline={Nanshan district},
	city={Shenzhen},
	postcode={518000},
	state={Guangdong},
	country={China}}

\ead{llshen@szu.edu.cn}


\author[6]{Shaohua Wan}

\affiliation[6]{organization={University
of Electronic Science and Technology of China},
	addressline={Longhua district},
	city={Shenzhen},
	postcode={518000},
	state={Guangdong},
	country={China}}
\ead{shaohua.wan@uestc.edu.cn}
  \author[1]{Jun Zhang}


\ead{zhangj@usx.edu.cn}  

\author[label1]{Yujing M Jiang} 

\affiliation[label1]{organization={United Aircraft AI Research Centre},
            country={Singapore}}


\begin{abstract}
Ensuring a stable power supply in rural areas relies heavily on effective inspection of power equipment, particularly transmission lines (TLs). However, detecting TLs from aerial imagery can be challenging when dealing with misalignments between visible light (RGB) and infrared (IR) images, as well as mismatched high- and low-level features in convolutional networks. To address these limitations, we propose a novel Hierarchical Multi-Modal Enhancement Network (HMMEN) that integrates RGB and IR data for robust and accurate TL detection. Our method introduces two key components: (1) a Mutual Multi-Modal Enhanced Block (MMEB), which fuses and enhances hierarchical RGB and IR feature maps in a coarse-to-fine manner, and (2) a Feature Alignment Block (FAB) that corrects misalignments between decoder outputs and IR feature maps by leveraging deformable convolutions. We employ MobileNet-based encoders for both RGB and IR inputs to accommodate edge-computing constraints and reduce computational overhead. Experimental results on diverse weather and lighting conditions—fog, night, snow, and daytime—demonstrate the superiority and robustness of our approach compared to state-of-the-art methods, resulting in fewer false positives, enhanced boundary delineation, and better overall detection performance. This framework thus shows promise for practical large-scale power line inspections with unmanned aerial vehicles.
\end{abstract}


\end{frontmatter}



\section{Introduction}
\label{sec:intro}

Industrial systems are increasingly deployed in rural regions, where a stable power supply is paramount for continuous and efficient operation. Within this context, transmission line inspection plays a pivotal yet demanding role. Conventionally, power utilities rely on telescopes and visual checks to spot defects in individual facilities, a process that is labor-intensive, time-consuming, and hazardous—particularly when lines extend over rivers, plains, and mountainous areas \cite{Choi2022attention}. These limitations underscore the urgent need for modern, automated detection strategies.
To address these challenges, advanced inspection platforms—such as Unmanned Aerial Vehicles (UAVs) \cite{zheng2021accurate}, hybrid robotic systems, and climbing robots—have been introduced to elevate both efficiency and safety. Equipped with diverse sensors (e.g., lidar, optical cameras, and thermal/infrared cameras), these platforms capture large volumes of aerial imagery processed by computer vision-based detection algorithms. Harnessing these data sources, an array of transmission line detection (TLD) methods have been developed \cite{pan2016power,song2014power,ha2017mfnet,oh20173Dpower,sun2019rtfnet,senthilnath2021bs,chen2023power,abdelfattah2022plgan,zhao2022amethod,yang2022ple}. Although some methods rely on RGB imagery \cite{pan2016power,song2014power,abdelfattah2022plgan,chen2023power,yang2022ple}, others focus on infrared (IR) images alone \cite{yang2022ple}, and still others combine RGB and IR data \cite{choi2019realtime,Choi2022attention}. While each of these paths has merit, single-sensor approaches can be hampered by environmental or resolution limitations, and naive fusion of both sensors often suffers from image misalignment.

Over the years, transmission line detection in computer vision has evolved along multiple lines of research, each leveraging different modalities. Early studies predominantly employed rule-based methods, relying on the thin, elongated geometry of transmission lines. These approaches employ edge detectors and thresholding to isolate potential line candidates \cite{song2014power,zhao2022amethod}, which are then refined through Markov random fields and least-squares fitting. As Convolutional Neural Networks (CNNs) gained traction, many methods pivoted to RGB-based TLD \cite{pan2016power,UNet2015,chen2018deeplab,yang2022ple,chen2023power}, exploiting the high resolution and color detail but still encountering challenges in adverse weather or low-light conditions.
Meanwhile, several researchers examined IR-based techniques \cite{yang2022ple,choi2019realtime} to overcome varying illumination. IR imagery is inherently robust to night-time or foggy scenarios, yet its lower resolution and narrower contrast range can result in suboptimal detection performance compared to color-based methods. 

Recognizing the complementary strengths of both RGB and IR data~\cite{chen2025acfnet,zhang2021image,wang2025mmae,zhang2025novel,li2025reference,liu2025three,tang2024itfuse,xing2024cfnet}, a growing body of work investigates multi-modal fusion \cite{choi2019realtime,Choi2022attention,ha2017mfnet,sun2019rtfnet,wang2016learning}, aiming to combine the detailed features from RGB with IR’s resilience to challenging environmental factors.
However, simply stacking RGB and IR inputs is often inadequate due to the inherent misalignment between the two modalities. This misalignment arises from differences in sensor resolutions, fields of view, and even slight positional shifts during data acquisition \cite{choi2019realtime}. Such discrepancies hinder the network’s ability to effectively fuse complementary information from RGB and IR images. Moreover, many U-Net-like architectures employed in TLD \cite{Choi2022attention,zhou2023transmission} amplify this problem by failing to account for feature-scale inconsistencies between high-level semantic features and low-level spatial details. These unresolved issues often result in suboptimal performance, particularly in scenarios requiring precise boundary delineation or robust detection under varying environmental conditions.
Addressing these limitations necessitates more refined alignment strategies and hierarchical integration techniques capable of reconciling feature misalignments across modalities and layers. The development of such methods is crucial for advancing multi-modal TLD into a robust and practical solution for real-world applications, where environmental variability and data diversity are unavoidable.

To tackle these challenges, this paper introduces an innovative approach to TLD, leveraging hierarchical multi-modal feature enhancement and alignment. The proposed method explicitly addresses the misalignments present in both multi-modal inputs and feature maps. Specifically, we design two key components: (1) the Mutual Multi-modal Enhanced Block (MMEB), which enhances the representational power of each modality by incorporating complementary information from the other, and (2) the Feature Alignment Block (FAB), which aligns feature maps from various sources, such as decoder outputs and IR feature maps, with those derived from RGB images. Together, these modules enable a more precise and robust fusion of RGB and IR data, ensuring superior detection accuracy and boundary preservation.

In summary, the key contributions of this work are as follows:
\begin{enumerate}
\item Hierarchical Feature Enhancement: We propose a novel MMEB to enhance the representational ability of RGB and IR feature maps by leveraging their complementary information in a hierarchical manner.  
\item We introduce the FAB to address feature map misalignments between modalities and across network layers, ensuring more coherent and accurate predictions.  
\item We conduct extensive experiments under diverse conditions, including varying weather and lighting, to validate the robustness of the proposed modules. Comparative experiments against state-of-the-art methods demonstrate the superior performance of our approach in terms of accuracy and robustness.  
\end{enumerate}

In summary, this work addresses the critical limitations of existing TLD approaches by tackling both input-level and feature-level misalignments, ensuring a more effective fusion of RGB and IR data. By leveraging hierarchical enhancements and alignment strategies, it lays the groundwork for multi-modal systems that are not only more robust and accurate but also adaptable to the diverse and challenging conditions encountered in real-world scenarios.

\section{Related Work}
\label{sec:rw}

The proposed method focuses on the automatic detection of transmission lines (TLs) using infrared (IR) and RGB images. This section reviews relevant works in the areas of transmission line detection (TLD), deep learning attention mechanisms, and feature fusion techniques, highlighting their contributions and limitations.

\subsection{Transmission Line Detection}

Transmission line detection (TLD) has garnered significant attention within the industrial research community due to its importance in power infrastructure monitoring. However, detecting transmission lines in real-world settings remains a challenging task. Transmission lines are thin and elongated, often blending into complex backgrounds, making them difficult to distinguish. Additionally, RGB images are sensitive to environmental conditions such as haze, snow, and rain, which can significantly degrade detection performance \cite{Choi2022attention,zhou2023transmission}.
Earlier methods for TLD were largely rule-based and relied on geometric characteristics of TLs, such as their elongated shapes. These methods used edge detection techniques (e.g., Canny operator, Otsu thresholding) and line-fitting algorithms (e.g., Hough transform) to identify TL candidates \cite{song2014power,zhao2022amethod}. While these approaches were effective in simple settings, they struggled with complex backgrounds and varying environmental conditions.

With the rapid advancement of Convolutional Neural Networks (CNNs), TLD methods evolved to leverage deep learning. Early CNN-based models first localized transmission line regions and then refined them using traditional line-detection methods \cite{pan2016power}. This approach improved accuracy but often failed to capture the fine boundaries of TLs due to the coarse nature of deep features. To address this, fully end-to-end segmentation networks such as U-Net \cite{UNet2015} and DeepLab \cite{chen2018deeplab} were adopted, enabling pixel-level predictions with improved precision. Recent advancements \cite{choi2019realtime,Choi2022attention,yang2022ple,chen2023power} have further improved TLD accuracy, yet challenges remain.
One major limitation of existing methods is their inability to fully address misalignments in feature representations. High-level semantic features from deeper layers often misalign with low-level spatial features, leading to blurred or inaccurate boundaries. Additionally, in multi-modal approaches that combine RGB and IR images, misalignments between the two modalities (e.g., differences in resolution, field of view) are frequently overlooked, reducing the potential benefits of data fusion.

To overcome these challenges, this work introduces a novel TLD method that utilizes RGB and IR images. The proposed method incorporates a feature enhancement block and a feature alignment block to address misalignments at both the input and feature levels, thereby improving detection accuracy and boundary delineation.

\subsection{Deep Learning Attention}

Attention mechanisms in deep learning are designed to highlight useful information while suppressing irrelevant or noisy features, making them particularly effective for tasks requiring fine-grained feature analysis. Among these, channel-wise attention has gained popularity in computer vision due to its simplicity and effectiveness.
The Squeeze-and-Excitation (SE) block \cite{Hu2018CVPR} is one of the most well-known attention mechanisms. It operates in two steps: (1) a squeeze operation that aggregates global information across spatial dimensions to produce channel-wise representative values, and (2) an excitation operation that re-weights the feature maps based on their importance. The SE block has been widely applied in various tasks, significantly enhancing performance by focusing on the most relevant features. To improve computational efficiency, Tang et al. \cite{tang2020nondestructive} replaced the fully connected (FC) layers in the SE block with 1×1 convolutions.

While these attention mechanisms are effective, they often focus exclusively on channel attention, ignoring spatial relationships within the feature maps. This limitation can lead to suboptimal results, particularly in tasks like TLD, where spatial information is critical for distinguishing the thin and elongated structures of transmission lines. To address this gap, researchers such as Choi et al. \cite{Choi2022attention} have integrated attention mechanisms into TLD systems to activate more useful features. However, these approaches still fail to capture the joint spatial and channel relationships effectively.

In this work, we propose a novel fusion method that integrates spatial and channel attention into a unified framework. This approach ensures that both spatial relationships and channel-level importance are captured, enabling more precise and robust feature enhancement. By addressing the limitations of existing attention mechanisms, the proposed method further strengthens the detection of transmission lines in complex and challenging environments.

\begin{figure*}[t]\footnotesize
	\begin{center}
		\includegraphics[width = 0.95\textwidth]{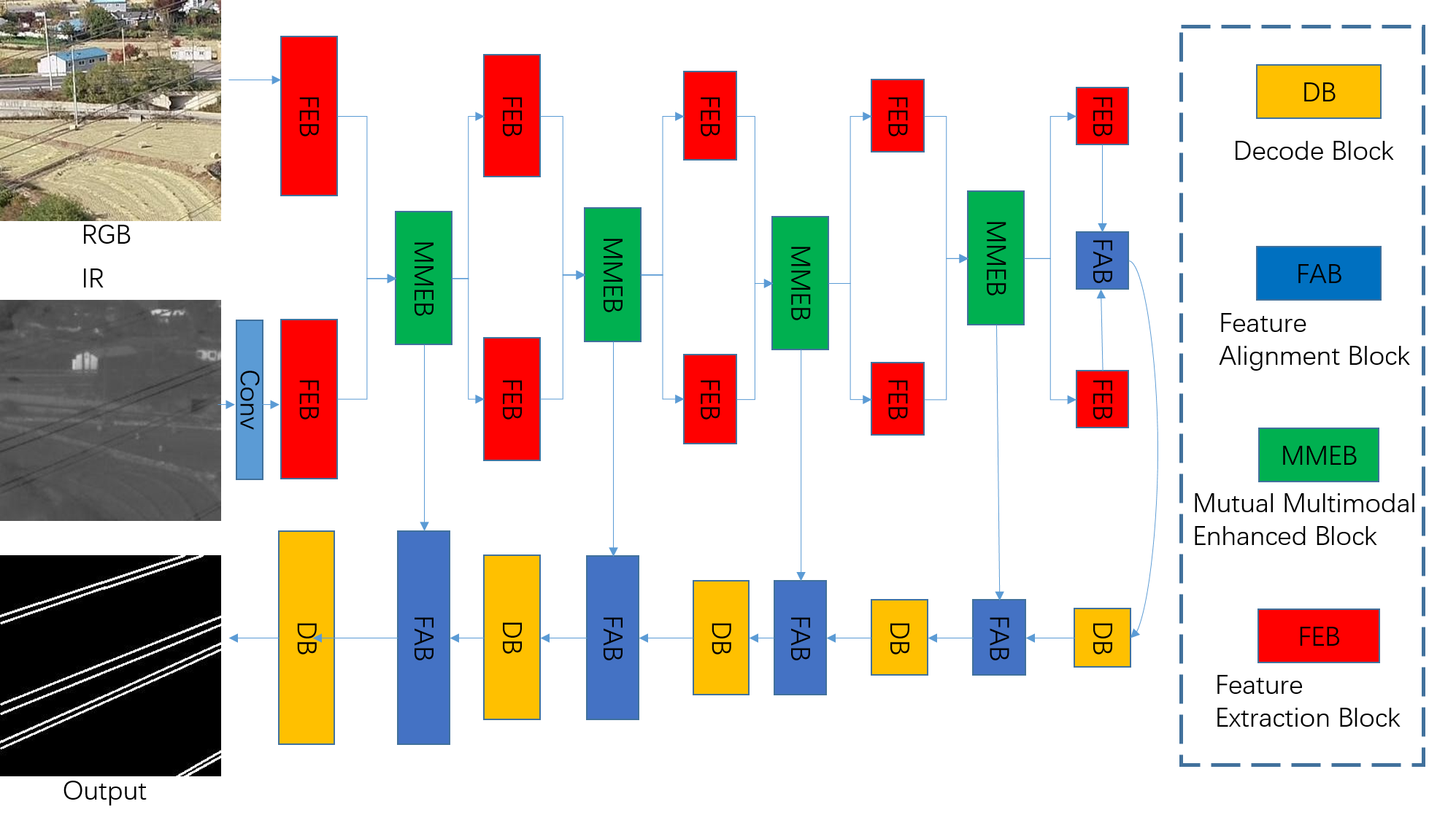}
	\end{center}
	\vspace{-2mm}
	\caption{Architecture of the proposed Hierarchical Mutual Multi-modal Enhanced Network (HMMEN). The proposed model consists of DB, MMEB, FAB, and DB. All outputted features from the MMEB are passed to the FAB at the same scale. The FAB at the lowest resolution only aligns the features from MMEB.
	}
		\vspace{-2mm}
	\label{fig-net}
\end{figure*}

\subsection{Feature Fusion}
Effective feature fusion~\cite{chen2025acfnet,zhang2021image} is essential for integrating multi-modal inputs, such as RGB and IR images, to exploit their complementary strengths. Traditional fusion methods aim to boost the quality of inputs and provides more information for follow-up tasks. In the context of transmission line detection (TLD), where environmental challenges and fine-grained features complicate the task, feature fusion methods play a critical role in achieving robust and accurate detection. Building on the advancements in multi-modal learning highlighted earlier, researchers have explored various strategies to combine features from different data sources while addressing their inherent disparities.

RTFNet \cite{sun2019rtfnet}, for example, fuses RGB and depth images for semantic segmentation tasks in indoor environments. By using encoder features to refine decoder-generated feature maps, it effectively integrates complementary information to enhance segmentation performance. This approach underscores the potential of leveraging multi-modal data for challenging tasks, such as TLD. Similarly, Choi et al. \cite{Choi2022attention} proposed a fusion module tailored to RGB and IR images, focusing on combining feature maps to create a unified representation for transmission line detection. While effective in certain scenarios, these methods often overlook misalignments between modalities, such as differences in sensor resolution, field of view, or image registration accuracy, which can hinder the effectiveness of the fused features.

Addressing such misalignments is crucial for ensuring the success of multi-modal systems, as poorly aligned features can lead to inconsistent or inaccurate predictions. Existing methods tend to rely on straightforward concatenation or simplistic fusion techniques, which fail to account for the varying contributions of each modality to the detection task. This limitation underscores the need for more sophisticated feature fusion strategies that dynamically adjust to the characteristics of the input data, ensuring coherent and complementary integration of multi-modal features.

\section{Hierarchical Multi-Modal TLD}
\label{sec:mt}

In this section, we introduce the motivation behind the proposed model and describe its architecture, as illustrated in Fig.~\ref{fig-net}. The model is designed to address the challenges of multi-modal transmission line detection (TLD), particularly misalignments between RGB and IR inputs and inconsistencies in feature representations across network layers. The architecture comprises four key components: the Feature Extract Block (FEB), Mutual Multi-modal Enhanced Block (MMEB), Feature Alignment Block (FAB), and Decode Block (DB).

The Feature Extract Block (FEB) is responsible for extracting hierarchical features from both RGB and IR images, laying the groundwork for subsequent feature enhancement and alignment. The Mutual Multi-modal Enhanced Block (MMEB) enhances these features by leveraging the complementary strengths of the two modalities, enabling more robust representations. The Feature Alignment Block (FAB) addresses feature misalignment issues by aligning the enhanced features from the MMEB with decoder-generated features, ensuring consistency across different feature scales. Finally, the Decode Block (DB) refines and processes the aligned features to produce high-resolution predictions of transmission line areas.

\begin{figure*}[!htbp]
	\begin{center}
		\tabcolsep 1pt
		\begin{tabular}{@{}ccccc@{}}
			\includegraphics[width = 0.194\textwidth]{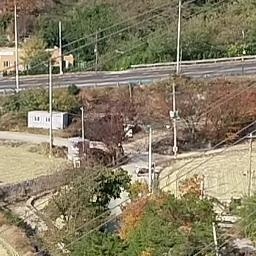} &
			\includegraphics[width = 0.194\textwidth]{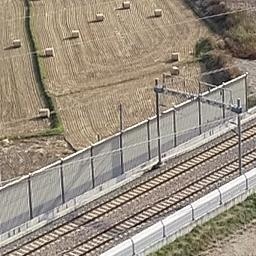}&
             \includegraphics[width = 0.194\textwidth]{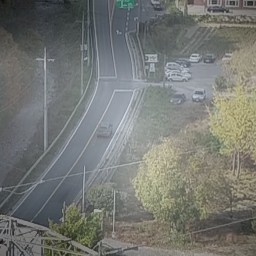} &
			\includegraphics[width = 0.194\textwidth]{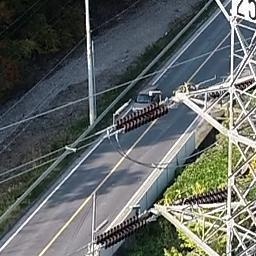}&
			\includegraphics[width = 0.194\textwidth]{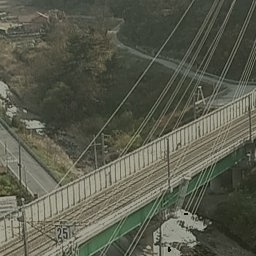}   \\
			
           \includegraphics[width = 0.194\textwidth]{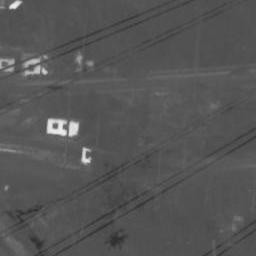} &
			\includegraphics[width = 0.194\textwidth]{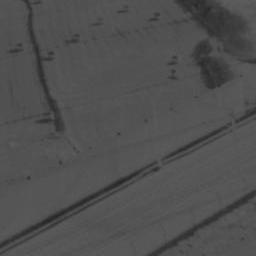}&
             \includegraphics[width = 0.194\textwidth]{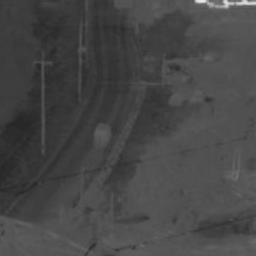} &
			\includegraphics[width = 0.194\textwidth]{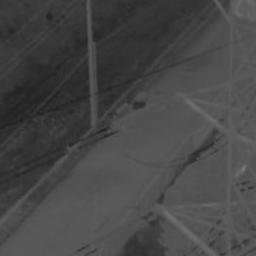}&
			\includegraphics[width = 0.194\textwidth]{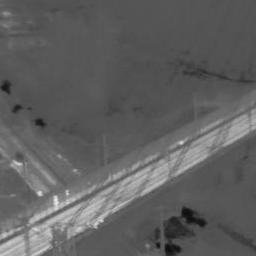}   \\

			 \includegraphics[width = 0.194\textwidth]{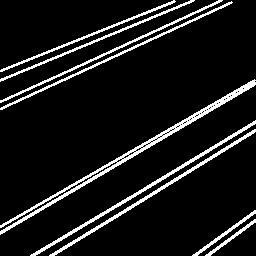} &
			\includegraphics[width = 0.194\textwidth]{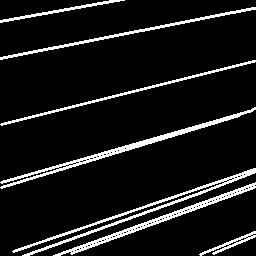}&
             \includegraphics[width = 0.194\textwidth]{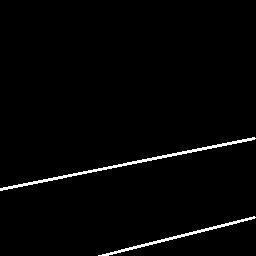} &
			\includegraphics[width = 0.194\textwidth]{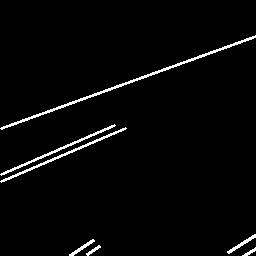}&
			\includegraphics[width = 0.194\textwidth]{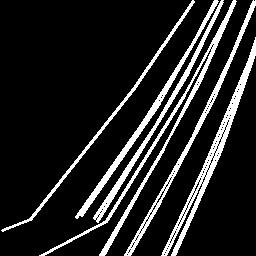}   \\
			(a)&(b)&(c)&(d)&(e)\\
		\end{tabular}
	\end{center}
	\vspace{-2mm}
	\caption{A visual example of matched RGB and IR images alongside their corresponding ground truths (GTs). The first row displays the RGB images, the second row shows the IR images, and the third row presents the GTs.}
\vspace{-3mm}
	\label{fig-demo}
\end{figure*}
\subsection{Motivations}

Modern UAVs are equipped with multiple sensors, yet previous TLD methods predominantly rely on RGB images alone, limiting their effectiveness in diverse environmental conditions. Single-sensor approaches face several challenges:
\begin{enumerate}
\item Sensitivity to Environmental Factors: Visible-light images are highly sensitive to illumination and weather conditions. For instance, as shown in Fig.~\ref{fig-demo}(c), RGB images often appear hazy in adverse weather, reducing their reliability.
\item Complex Backgrounds: RGB images frequently include intricate backgrounds that resemble transmission lines, as seen in Fig.~\ref{fig-demo}(a) and (b), making it harder to accurately isolate the lines.
\item Low Contrast in IR Images: While IR images are robust to lighting variations, they suffer from low resolution and poor contrast between transmission lines and the background, as illustrated in Fig.~\ref{fig-demo}(d) and (e).
\end{enumerate}

Despite these individual limitations, RGB and IR images offer complementary strengths. RGB images provide high-resolution and detailed visual information essential for identifying transmission lines, while IR images demonstrate resilience to weather and lighting variability, mitigating the impact of challenging environmental conditions. By combining these modalities, the proposed architecture is designed to extract discriminative and robust features from both inputs, improving detection performance.

Capturing large contextual information is essential for accurate transmission line detection (TLD), particularly for delineating the fine, elongated structures of transmission lines. CNN models achieve this through downsampling, which expands the receptive field and captures global context, followed by upsampling to restore feature map sizes for dense predictions. However, integrating high-level semantic features from upsampled maps with low-level spatial features often results in misalignment due to differences in resolution and scale, leading to blurred or inaccurate predictions.
This misalignment poses a significant challenge, especially in TLD, where transmission lines occupy a small portion of the image and often blend into complex backgrounds. To address this, the proposed Feature Alignment Block (FAB) ensures coherent integration of upsampled high-level features with high-resolution low-level features. By dynamically aligning these features, the FAB preserves both global context and local detail, enabling the model to distinguish transmission lines more effectively.
Together with the Mutual Multi-modal Enhanced Block (MMEB), which strengthens feature representations from RGB and IR inputs, the FAB forms a cohesive architecture that resolves feature misalignments and enhances multi-modal integration. This design ensures robust and precise transmission line detection, even in challenging real-world environments.

\subsection{Mutual Multi-modal Enhanced Block (MMEB)}

\begin{figure*}[t]\footnotesize
	\begin{center}
		\includegraphics[width = 0.99\textwidth]{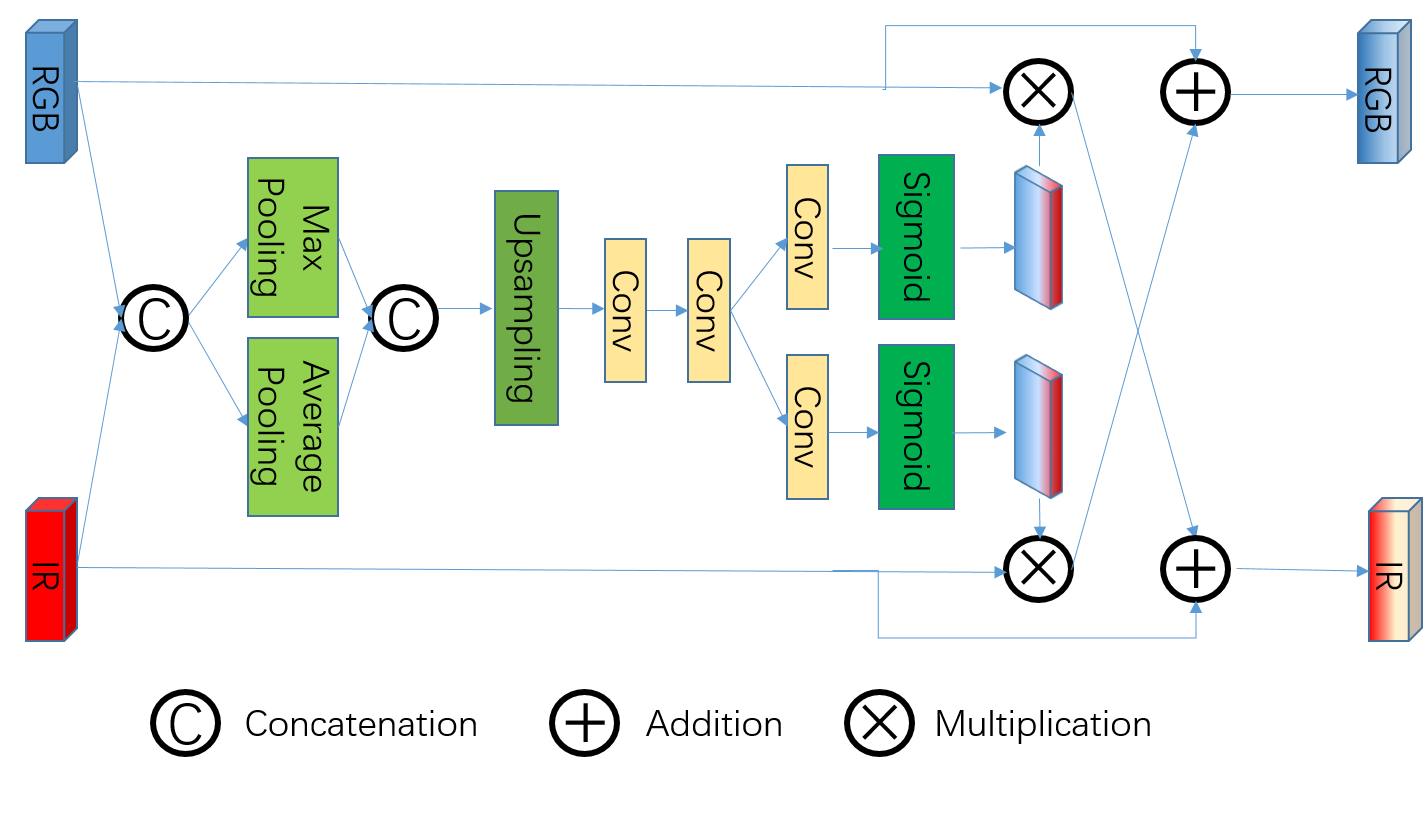}
	\end{center}
	\vspace{-3mm}
	\caption{Architecture of the proposed Mutual Multi-modal Enhanced Block (MMEB), illustrating the process of feature extraction, weight map calculation, and mutual enhancement for RGB and IR modalities.}
		\vspace{-2mm}
	\label{fig-ffb}
\end{figure*}

Effectively fusing RGB and IR images for transmission line detection (TLD) remains challenging despite their complementary strengths. RGB and IR images suffer from spatial misalignment, often caused by differences in resolution, field of view, or inaccuracies in registration techniques such as MATLAB-based image matching \cite{Choi2022attention}. Additionally, RGB and IR modalities contribute unevenly to TLD, with their strengths varying depending on the environmental conditions and image content. To address these challenges, we propose the Mutual Multi-modal Enhanced Block (MMEB), which enhances the representational capacity of each modality by leveraging complementary information from the other.

As illustrated in Fig.~\ref{fig-ffb}, the MMEB receives feature maps from RGB and IR images as input and outputs refined feature maps for the next encoder level and corresponding decoder block. The process involves the following steps:
\begin{enumerate}
    \item Feature Concatenation: Feature maps from RGB and IR images are concatenated to form a unified feature map, $\boldsymbol{F}_1$.
    \item Spatial Feature Extraction: Max-pooling and average-pooling operations are applied to $\boldsymbol{F}_1$ to generate a spatially representative feature map, $\boldsymbol{F}_2$.
    \item Weight Map Calculation: The feature map $\boldsymbol{F}_2$ is upsampled and processed through a series of convolution layers followed by sigmoid activations to generate weight maps $\boldsymbol{W}_{\text{RGB}}$ and $\boldsymbol{W}_{\text{IR}}$ for the RGB and IR modalities, respectively.
    \item Feature Enhancement: The enhanced feature maps are calculated using the equations:
   \begin{equation}
   \boldsymbol{EF}_{\text{RGB}} = F_{\text{RGB}} + F_{\text{IR}} \cdot \boldsymbol{W}_{\text{IR}},
   \label{eq-efrgb}
   \end{equation}
   \begin{equation}
   \boldsymbol{EF}_{\text{IR}} = F_{\text{IR}} + F_{\text{RGB}} \cdot \boldsymbol{W}_{\text{RGB}}.
   \label{eq-efir}
   \end{equation}

\end{enumerate}

The resulting enhanced feature maps, $\boldsymbol{EF}_{\text{RGB}}$ and $\boldsymbol{EF}_{\text{IR}}$, are passed to the next level of the encoder and decoder, ensuring that both modalities mutually reinforce their strengths for improved detection performance.

\subsection{Feature Alignment Block (FAB)}

\begin{figure*}[t]\footnotesize
	\begin{center}
		\includegraphics[width = 0.99\textwidth]{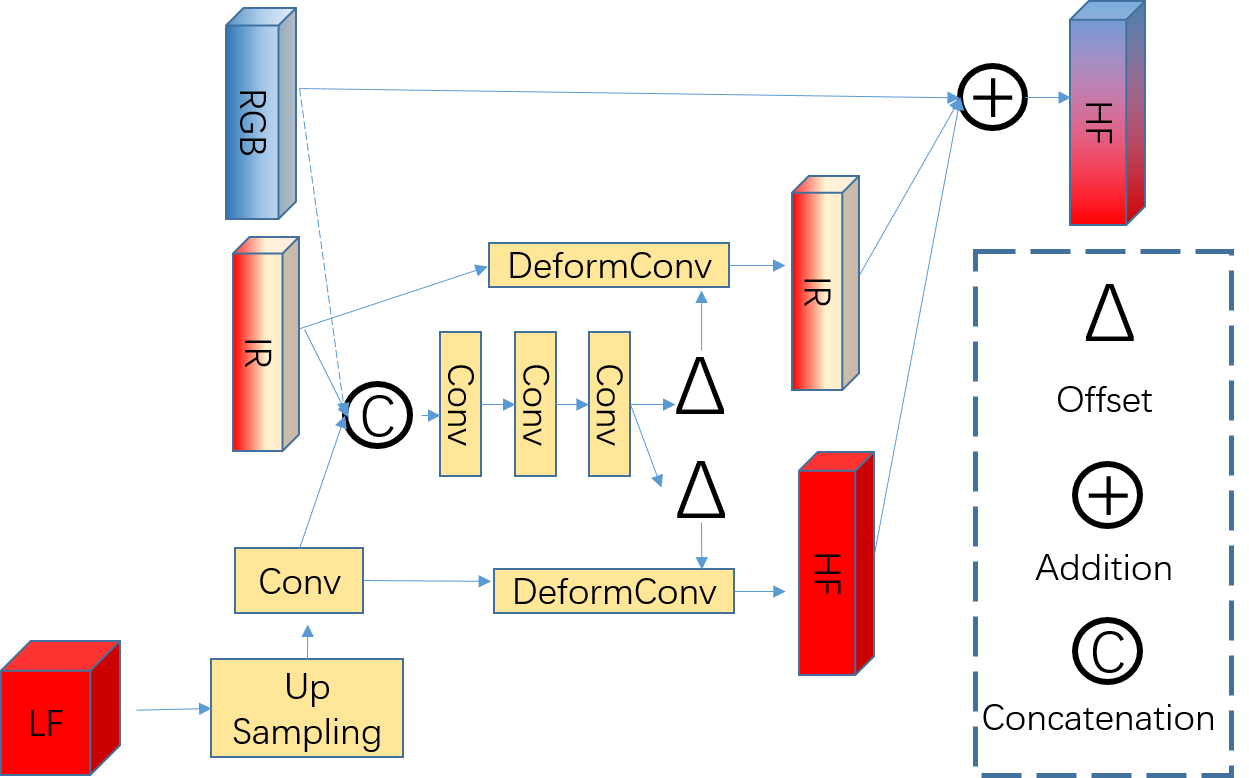}
	\end{center}
	\vspace{-2mm}
	\caption{Architecture of the proposed Feature Alignment Block (FAB), detailing the alignment process for upsampled high-level features and high-resolution low-level features to ensure coherent feature integration.
	}
		\vspace{-2mm}
	\label{fig-fab}
\end{figure*}

Large-scale contextual information is crucial for TLD, as it provides rich semantic guidance for accurate predictions. However, integrating high-level semantic features with high-resolution low-level features often results in misalignment, leading to degraded detection accuracy. To address this issue, we propose the Feature Alignment Block (FAB), which dynamically aligns feature maps across modalities and resolutions.

The architecture of FAB, shown in Fig.~\ref{fig-fab}, takes feature maps from the decoder block (DB) and feature extract block (FEB) as inputs. The process is as follows:
\begin{enumerate}
\item Upsampling: Low-resolution feature maps are upsampled to match the size of the high-resolution maps.
\item Channel Reduction: A convolution operation reduces the number of channels in the upsampled feature map, generating $F_{\text{llc}}$.
\item Offset Prediction: $F_{\text{llc}}$ is concatenated with the corresponding feature maps from RGB and IR inputs. Convolution layers are applied to predict offsets $\Delta_{\text{IR}}$ and $\Delta_{\text{llc}}$.
\item Feature Alignment: Deformable convolutions align the feature maps:
   \begin{equation}
   F_{\text{AUF}} = \text{Deform}(F_{\text{llc}}, \Delta_{\text{llc}}),
   \label{eq-auf}
   \end{equation}
   \begin{equation}
   F_{\text{AIF}} = \text{Deform}(F_{\text{IR}}, \Delta_{\text{IR}}),
   \label{eq-aif}
   \end{equation}
   where $F_{\text{AUF}}$ and $F_{\text{AIF}}$ are aligned feature maps for the upsampled high-level and IR feature maps, respectively.

\item Final Alignment: The final aligned feature map is obtained by merging all aligned feature maps:
   \begin{equation}
   F_{\text{Align}} = F_{\text{RGB}} + F_{\text{AUF}} + F_{\text{AIF}}.
   \label{eq-aif}
   \end{equation}

The FAB ensures that both high-level and low-level features are accurately aligned, preserving structural details crucial for transmission line detection.

\end{enumerate}

\subsection{Training Losses}

The training losses are carefully designed to ensure the proposed network effectively learns to detect transmission lines (TLs) in challenging scenarios. As transmission line detection (TLD) belongs to the category of binary pixel-level segmentation problems, appropriate loss functions are crucial for optimizing both pixel-wise accuracy and overall detection performance.

\subsubsection{Binary Cross-Entropy Loss}

The binary cross-entropy (BCE) loss is commonly used for binary segmentation tasks. It measures the discrepancy between the predicted probability ($p_i$) and the ground truth label ($g_i$) for each pixel. For TLD, the BCE loss is defined as:

\begin{equation}
\mathcal{L}_{B}=  -\frac{1}{N}\sum_{i=1}^{N}{\left[g_i \log p_i + (1-g_i) \log (1-p_i)\right]},
\label{eq-bloss}
\end{equation}

where \( N \) is the total number of pixels in the image. This loss ensures the model learns to minimize errors in predicting the binary classification (transmission line vs. background) for each pixel.

\subsection{Addressing Class Imbalance}

In TLD, a significant challenge arises from the class imbalance between transmission line pixels and background pixels. Transmission lines typically occupy a small fraction of an image, leading to an overwhelming majority of background pixels. This imbalance can bias the model toward predicting the dominant background class, thereby reducing detection performance for transmission lines. While the BCE loss captures pixel-wise discrepancies effectively, it does not account for this imbalance.

\subsection{Dice Loss}

To address the class imbalance problem, we incorporate the Dice loss~\cite{lyu2022attention}, which evaluates the overlap between the predicted segmentation and the ground truth. Dice loss directly optimizes for similarity between the predicted transmission line regions and the ground truth by emphasizing the relative importance of smaller, underrepresented classes like transmission lines. It is defined as:

\begin{equation}
\mathcal{L}_{d}= 1 - \frac{2\sum_{i=1}^{N} g_i p_i}{\sum_{i=1}^{N} g_i^2 + \sum_{i=1}^{N} p_i^2},
\label{eq-sloss}
\end{equation}

where \( g_i \) represents the ground truth binary label, and \( p_i \) is the predicted probability for pixel \( i \). Dice loss ensures the model prioritizes the accurate segmentation of transmission lines, mitigating the impact of class imbalance and reducing the risk of overfitting to background pixels.

\subsection{Final Loss Function}

To balance the strengths of the BCE and Dice losses, the total loss function combines the two, ensuring both pixel-wise accuracy and robust segmentation of transmission lines:

\begin{equation}
\mathcal{L}_{\text{total}}= \mathcal{L}_{B} + \lambda \mathcal{L}_{d},
\label{eq-tloss}
\end{equation}

where \( \lambda \) is a weighting factor that adjusts the relative importance of the Dice loss. This combination allows the model to learn effectively, balancing the need for accurate classification of both dominant background pixels and sparsely distributed transmission line pixels.

By incorporating both BCE and Dice losses, the proposed loss function ensures robust optimization, enabling the model to handle the unique challenges of TLD, including class imbalance and the need for precise boundary delineation.

\section{Experimental Results}

In this section, we detail the experimental settings, including the dataset, evaluation metrics, baseline models, comparison results, and ablation studies.

\subsection{Transmission Line Dataset}

CNN-based models benefit significantly from access to large-scale datasets. For training and testing our proposed model, we utilized the RGB and IR transmission line detection dataset introduced by Choi et al.~\cite{Choi2022attention}, referred to as the VITLD dataset. This dataset was captured using a custom drone system equipped with a DJI Phantom, integrating both visible light and infrared cameras. The visible light camera captures high-resolution images ($1920 \times 1080$), while the infrared camera provides lower-resolution images ($640 \times 512$). 

To ensure diversity and represent real-world complexity, videos were recorded in varied settings, such as roads, groves, and railway tracks. Representative frames with varying contexts were then selected to reduce redundancy and increase dataset variety. However, an initial mismatch between RGB and IR images was observed due to differences in the cameras' fields of view and resolutions. To address this, Choi et al.~\cite{Choi2022attention} employed MATLAB's image registration tool to align the images. Following alignment, the images were cropped and resized to a consistent dimension of $256 \times 256$. The final dataset comprised 400 matched RGB-IR image pairs. For further technical details on the dataset construction, we refer readers to~\cite{Choi2022attention}.

The primary application of transmission line detection (TLD) is in outdoor environments, where low light and adverse weather conditions pose significant challenges. To simulate such scenarios and enhance the robustness of the model, Choi et al.~\cite{Choi2022attention} employed data augmentation techniques. RGB images were augmented using the Python library \textit{imageaug} to simulate various environmental conditions, including fog, snow, night, and daytime scenarios. Meanwhile, infrared (IR) images, known for their robustness to environmental variations, were retained in their original form as counterparts to the augmented RGB images. This augmentation strategy ensured that the dataset closely represented the challenges encountered in real-world TLD applications. For detailed information about the augmentation techniques, readers can refer to~\cite{Choi2022attention}.

\begin{figure*}[!htbp]\scriptsize
	\centering
	\tabcolsep 0.5pt
	\begin{tabular}{ccccccccc}
\includegraphics[width=.11\linewidth]{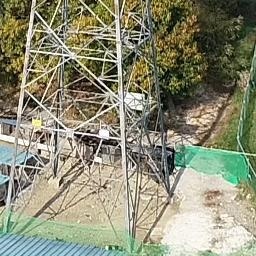}	&\includegraphics[width=.11\linewidth]{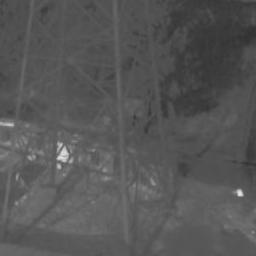}		
 &\includegraphics[width=.11\linewidth]{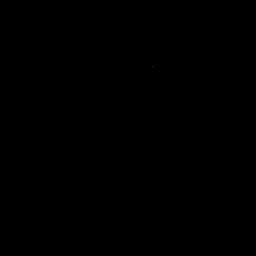}&\includegraphics[width=.11\linewidth]{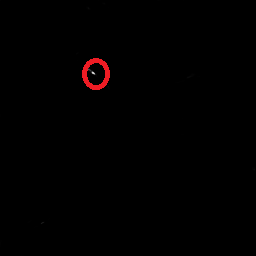}
&\includegraphics[width=.11\linewidth]{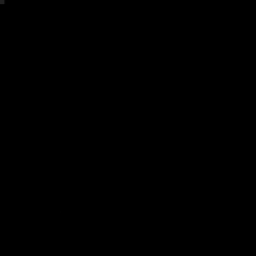}&\includegraphics[width=.11\linewidth]{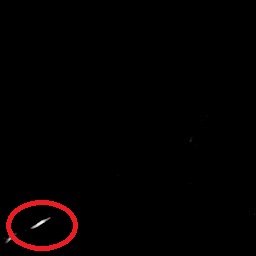}
&\includegraphics[width=.11\linewidth]{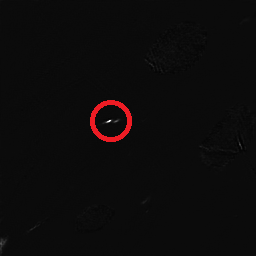}
&\includegraphics[width=.11\linewidth]{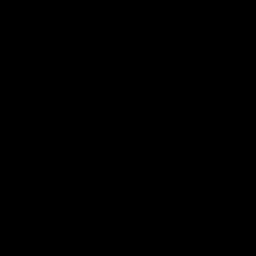}&\includegraphics[width=.11\linewidth]{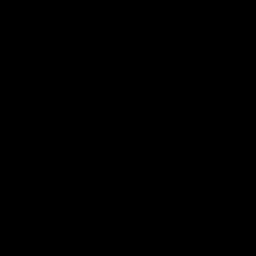}
							\\	

 \includegraphics[width=.11\linewidth]{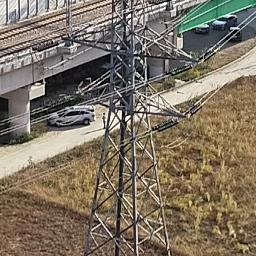}	&\includegraphics[width=.11\linewidth]{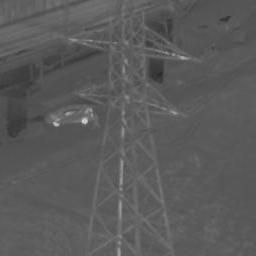}		
 &\includegraphics[width=.11\linewidth]{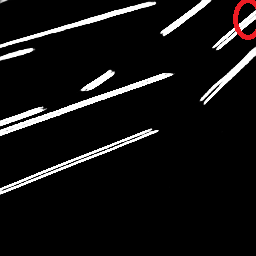}&\includegraphics[width=.11\linewidth]{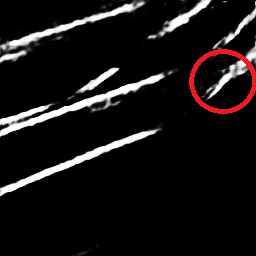}
&\includegraphics[width=.11\linewidth]{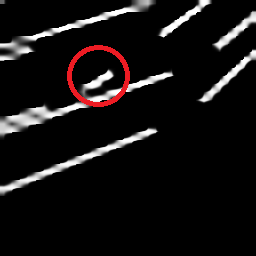}&\includegraphics[width=.11\linewidth]{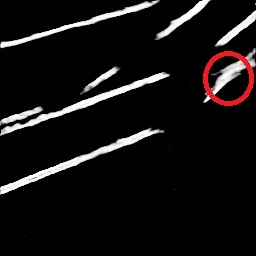}
&\includegraphics[width=.11\linewidth]{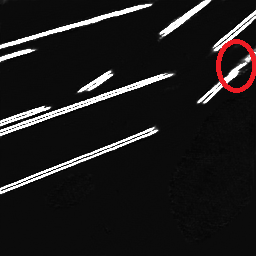}
&\includegraphics[width=.11\linewidth]{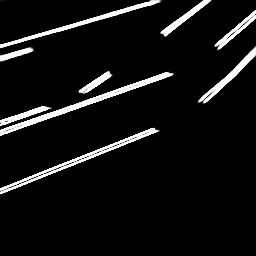}&\includegraphics[width=.11\linewidth]{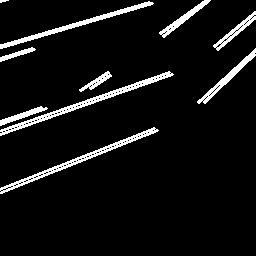}
							\\		
     \includegraphics[width=.11\linewidth]{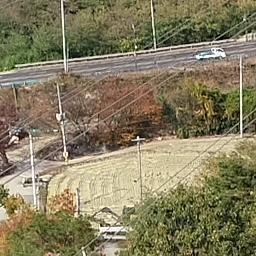}	&\includegraphics[width=.11\linewidth]{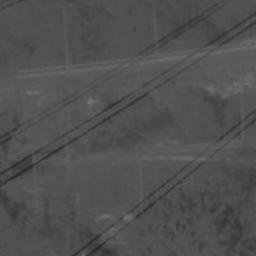}		
 &\includegraphics[width=.11\linewidth]{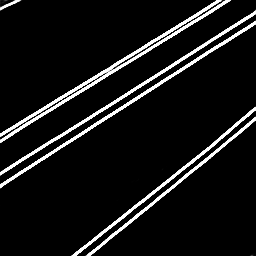}&\includegraphics[width=.11\linewidth]{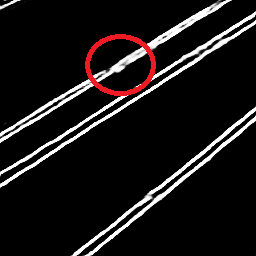}
&\includegraphics[width=.11\linewidth]{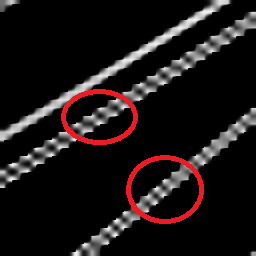}&\includegraphics[width=.11\linewidth]{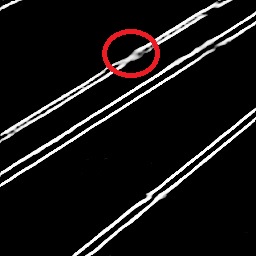}&
\includegraphics[width=.11\linewidth]{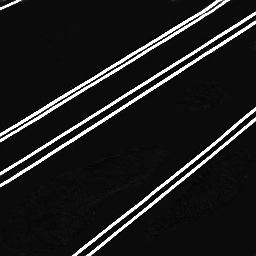}
&\includegraphics[width=.11\linewidth]{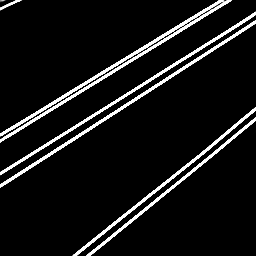}&\includegraphics[width=.11\linewidth]{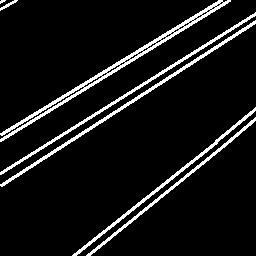}
							\\
\includegraphics[width=.11\linewidth]{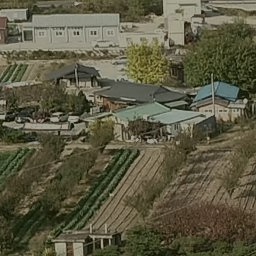}	&\includegraphics[width=.11\linewidth]{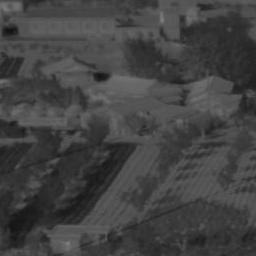}		
 &\includegraphics[width=.11\linewidth]{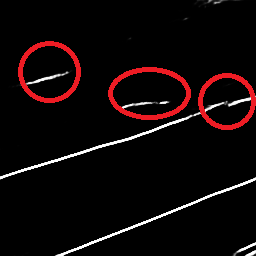}&\includegraphics[width=.11\linewidth]{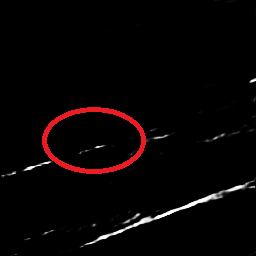}
&\includegraphics[width=.11\linewidth]{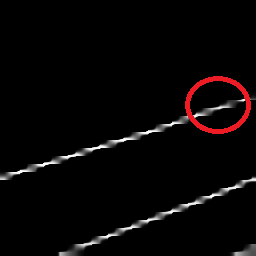}&\includegraphics[width=.11\linewidth]{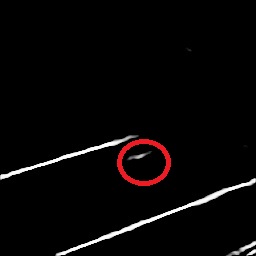}
&\includegraphics[width=.11\linewidth]{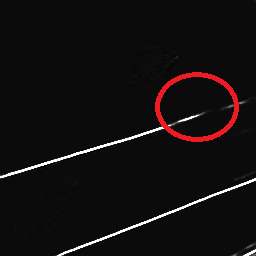}
&\includegraphics[width=.11\linewidth]{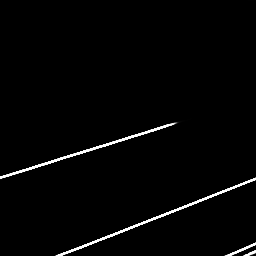}&\includegraphics[width=.11\linewidth]{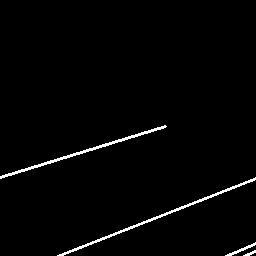}
							\\	
\includegraphics[width=.11\linewidth]{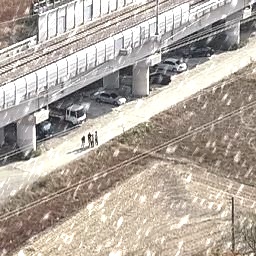}	&\includegraphics[width=.11\linewidth]{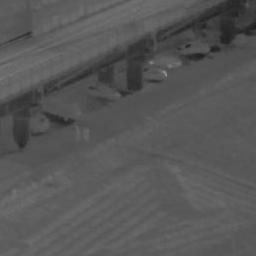}		
 &\includegraphics[width=.11\linewidth]{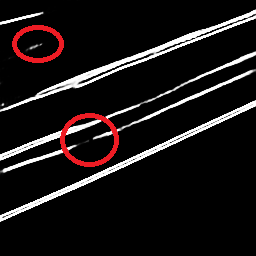}&\includegraphics[width=.11\linewidth]{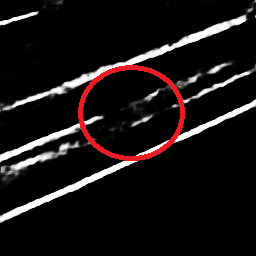}
&\includegraphics[width=.11\linewidth]{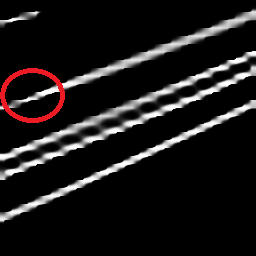}&\includegraphics[width=.11\linewidth]{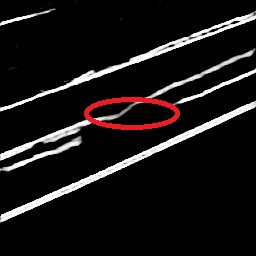}
&\includegraphics[width=.11\linewidth]{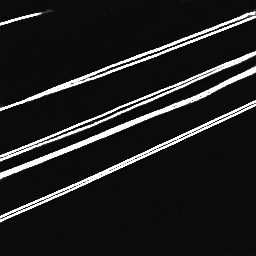}
&\includegraphics[width=.11\linewidth]{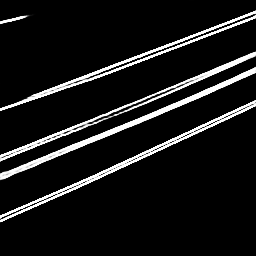}&\includegraphics[width=.11\linewidth]{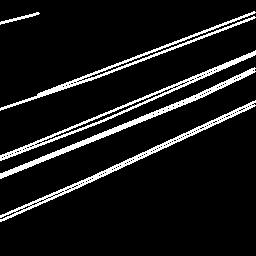}
							\\	
	
\includegraphics[width=.11\linewidth]{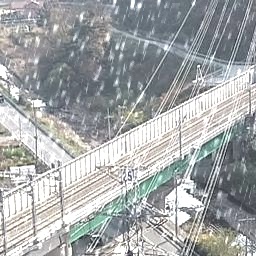}	&\includegraphics[width=.11\linewidth]{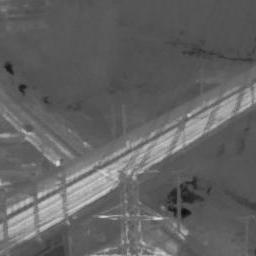}		
 &\includegraphics[width=.11\linewidth]{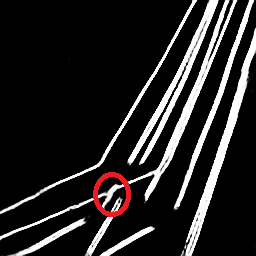}&\includegraphics[width=.11\linewidth]{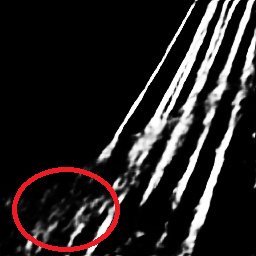}
&\includegraphics[width=.11\linewidth]{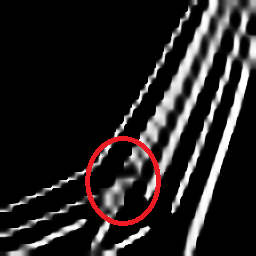}&\includegraphics[width=.11\linewidth]{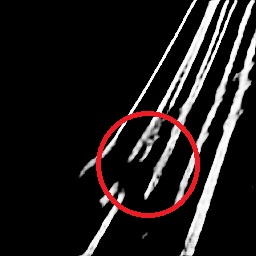}
&\includegraphics[width=.11\linewidth]{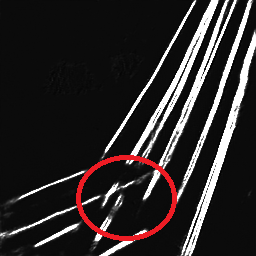}
&\includegraphics[width=.11\linewidth]{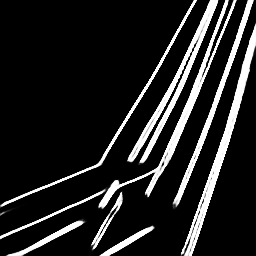}&\includegraphics[width=.11\linewidth]{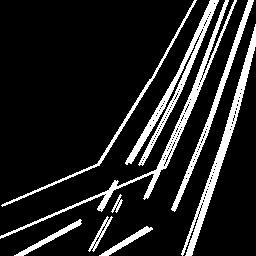}
							
							\\
		(a) RGB & (b) IR & (c) SegNet & (d) LHRNet & (e) Bisenet&(e) BGRNet & (f) MFNet &(g) HMFN  &(h) GTs\\
		
	\end{tabular}
	
	\caption{Visual examples of transmission line (TL) detection results using various methods. Areas circled in red highlight key differences and details for comparison.}
	\label{fig:comresult}
\vspace{-4mm}
\end{figure*}

\subsection{Implementation Details}

The proposed model is built upon the U-Net architecture, a widely used framework for image segmentation tasks~\cite{UNet2015,Choi2022attention,yang2022ple}. Recognizing the need for efficient computation on IoT devices, we chose MobileNet~\cite{sandler2018mobilenetv2} as the backbone for feature extraction from RGB and IR images. MobileNet offers a lightweight and resource-efficient alternative to heavier architectures like DenseNet~\cite{huang2017densely} and ResNet~\cite{he2016deep}, making it more suitable for edge applications. 

In the MobileNet-based U-Net, feature maps of the same resolution are grouped into Feature Extract Blocks (FEBs) for hierarchical feature extraction. For the decoder block (DB), we adopt ConvTranspose operations and Inverted Residuals and Linear Bottlenecks~\cite{sandler2018mobilenetv2} to upsample and refine the feature maps. These components ensure the model captures both global context and local details while maintaining computational efficiency. For detailed implementation of these operations, refer to~\cite{sandler2018mobilenetv2}.

The model is implemented using the PyTorch deep learning framework. Although designed for lightweight computation, the model was trained on a machine with 8 TITAN V GPUs, with the capability to run effectively on a single GPU. The Adam optimizer was employed with an initial learning rate of 0.0001, which decayed by a factor of 0.5 every 25 epochs after the first 150 epochs. A batch size of five was used consistently during training, and other parameters were set to default values. 

The dataset introduced in~\cite{Choi2022attention} was divided into three subsets: 70\% for training, 10\% for validation, and 20\% for testing. To enhance the model's robustness, data augmentation techniques, such as random horizontal flipping, random brightness adjustments, and random contrast variations, were applied during training. In the testing phase, the dataset was further augmented with simulated variations in weather and lighting conditions to evaluate the model's performance under diverse real-world scenarios.

\subsection{Evaluation Metrics}

To comprehensively assess the proposed model's performance, we employed seven metrics commonly used in image segmentation and TLD tasks~\cite{Choi2022attention,UNet2015,ha2017mfnet,badrinarayanan2017segnet,yu2018bisenet,yu2021lite}:

\begin{itemize}
    \item {Intersection over Union (IoU)}: A widely used pixel-level metric that measures the overlap between predicted and ground truth regions.
    \item {Pixel Accuracy (PA)}: Evaluates the percentage of correctly classified pixels over the total number of pixels.
    \item {Area under Precision-Recall Curve (AUC)}: Complements PA by summarizing the trade-off between precision and recall across different thresholds.
    \item {Precision}: Quantifies the proportion of correctly predicted positive pixels relative to all predicted positive pixels.
    \item {Sensitivity (Recall)}: Measures the percentage of true positive pixels correctly identified by the model.
    \item {Dice Coefficient}: Evaluates the similarity between the predicted and ground truth regions, with a focus on the overlap of positive regions.
\end{itemize}

In industrial applications, detecting all transmission lines is more critical than avoiding false positives. To address this, an object-level metric can be employed, which calculates the recall rate of transmission lines. A predicted region is considered a correctly detected transmission line if its overlap ratio with the ground truth exceeds a predefined threshold (e.g., 0.5). This metric ensures that the model prioritizes recall, minimizing the risk of missed detections in real-world scenarios.

\subsection{Baseline Models}

The proposed modules in this paper are highly adaptable and can be integrated into various semantic segmentation models. U-Net, a well-established semantic segmentation framework, has been widely adopted due to its effectiveness and ability to produce high-quality results. Following the methodology of~\cite{Choi2022attention}, our proposed model is built upon U-Net, which serves as the baseline model. U-Net’s encoder-decoder architecture provides a strong foundation for exploring and enhancing fusion and alignment modules, which are the primary focus of this paper, rather than the overarching model architecture.

To enable multi-modal input, we extended the original U-Net architecture—initially designed for single-modality data, particularly in medical image segmentation—by incorporating a second encoder for infrared (IR) images. As illustrated in Fig.~\ref{fig-net}, the original encoder processes RGB images, while the added encoder is dedicated to IR data. Since IR images are typically single-channel, we applied a convolution operation to convert them into three-channel feature maps, making them compatible with the network structure. These feature maps are then fed into the second encoder, ensuring efficient processing of both modalities.

This architecture allows us to demonstrate the flexibility and efficacy of the proposed modules by integrating them into a widely used framework. The enhanced U-Net, with its dual-encoder design, effectively combines features from RGB and IR inputs to improve transmission line detection performance. This dual-modality approach emphasizes the potential of the proposed modules for broader applications in multi-modal semantic segmentation tasks.

\subsection{Qualitative and Quantitative Comparison Results}

\begin{table}[h] \small
	\caption{Quantitative detection results of state-of-the-art methods and the proposed method on RGB and IR transmission line (TL) images. The highest scores are highlighted in bold. Results are represented as percentages for compactness.}
    \vspace{5mm}
	\label{Tab:com}
	\centering
	\begin{tabular}{c|c|c|c|c|c|c}
		\hline
	Method& $S_e$ & Dice & AUC & AP& $P_r$ & IoU\\
            \hline
          U-Net &64.0 &65.5& 80.8&98.4&67.7&56.9
		\\
            \hline
           SegNet&53.5 &56.4& 70.5&97.6&61.4&46.4
            		\\
            \hline

            MFNet&60.7&64.8&78.5&98.3&69.0&56.2
            	\\
            \hline
            LHRNet&52.3&52.56&66.7&97.1&54.4&41.6
            	\\
            \hline
           Bisenet&  47.6& 53.8 & 61.3&96.4&49.7&41.2
            		\\
            \hline
            BGRNet&68.1&69.4&83.3&98.7&71.3&62.3
            		\\
            \hline
		HMMEN& \textbf{72.3}& \textbf{72.6} & \textbf{86.2} &\textbf{99.1}&\textbf{73.0}&\textbf{66.6}		\\
		\hline
	\end{tabular}
\end{table}

In this study, we trained several state-of-the-art end-to-end segmentation methods, including MFNet \cite{ha2017mfnet}, SegNet \cite{badrinarayanan2017segnet}, BiSeNet \cite{yu2018bisenet}, Lite-HRNet \cite{yu2021lite}, and BGRNet \cite{zhou2023transmission}, on the TLD dataset~\cite{Choi2022attention} and compared their performance with our proposed method. To improve the detection capability of Lite-HRNet, we introduced modifications to its architecture. Specifically, we concatenated low-level features from the stem with the original Lite-HRNet output to create a new feature map \( F_c \). This feature map was then upsampled to the original input size and used for transmission line prediction. In this section, we refer to this modified Lite-HRNet as LHRNet.

Table~\ref{Tab:com} presents the performance of these segmentation models. Among the methods evaluated, SegNet, LHRNet, and BiSeNet exhibited the lowest detection scores. These models rely heavily on low-resolution feature maps that are subsequently upsampled to the original input size, thereby neglecting the critical low-level features extracted by the backbone. This oversight compromises their ability to accurately locate and delineate transmission lines. 
MFNet outperformed these models by leveraging high-resolution low-level features to refine the decoded feature maps, which is crucial for capturing the precise shape and location of transmission lines. However, while MFNet showed improvements, it was still surpassed by BGRNet, introduced by Zhou et al. \cite{zhou2023transmission}, which achieved the second-highest detection performance. BGRNet integrates multimodal data more effectively but is hindered by limitations in addressing feature misalignment.
Our proposed method demonstrated the highest detection accuracy among all evaluated models, attributed to its hierarchical multi-modal enhancement and feature alignment strategies. The hierarchical multi-modal enhancement module facilitates information exchange between RGB and IR features in a coarse-to-fine manner, enabling better feature representation. Meanwhile, the feature alignment block resolves misalignment issues between feature maps from different modalities and hierarchical levels, ensuring seamless integration of useful high-level features into the decoding process.

To further substantiate the superiority of our method, we provide qualitative comparisons in Fig.~\ref{fig:comresult}, which illustrate the performance of different methods under various challenging weather conditions, such as fog, snow, and low light. False detections and missed predictions are marked with red circles. These visual examples highlight the limitations of existing methods. For instance, LHRNet often fails to capture the precise shape of transmission lines, while SegNet frequently leaves parts of transmission lines undetected. Similarly, BGRNet, though effective in simpler scenarios, struggles under complex conditions, leading to a significant number of missed detections. In contrast, our method consistently achieves accurate line detection while minimizing false positives, even in challenging environments.
These results underscore that TLD is a unique problem requiring specialized approaches. Traditional semantic segmentation methods are insufficient to handle the intricacies of TLD effectively. By addressing the specific challenges of multimodal integration and feature alignment, our proposed method demonstrates substantial improvements in both detection accuracy and robustness.

\begin{figure*}[!htbp]\scriptsize
	\centering
	\tabcolsep 0.5pt
	\begin{tabular}{ccccccc}
 \includegraphics[width=.14\linewidth]{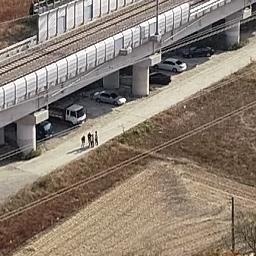}	&\includegraphics[width=.14\linewidth]{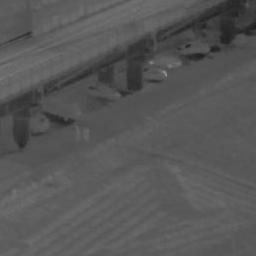}		
 &\includegraphics[width=.14\linewidth]{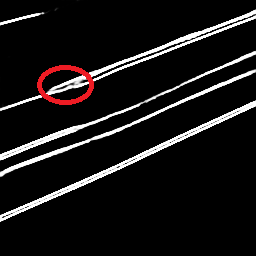}&\includegraphics[width=.14\linewidth]{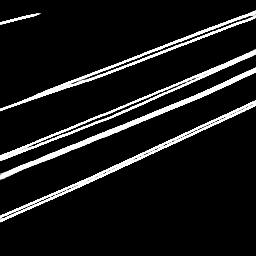}
&\includegraphics[width=.14\linewidth]{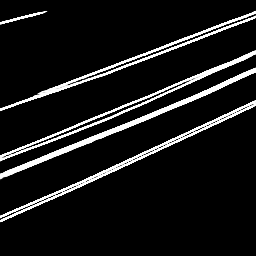}&\includegraphics[width=.14\linewidth]{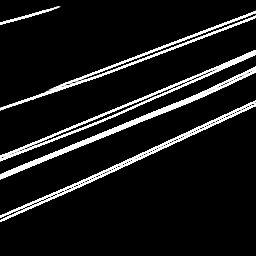}
&\includegraphics[width=.14\linewidth]{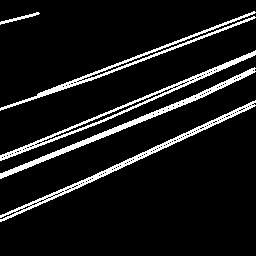}

							\\		
 \includegraphics[width=.14\linewidth]{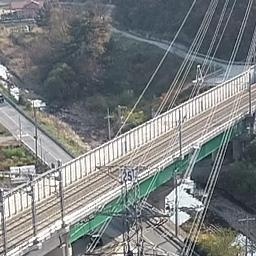}	&\includegraphics[width=.14\linewidth]{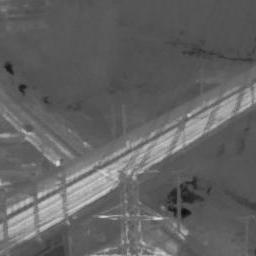}		
 &\includegraphics[width=.14\linewidth]{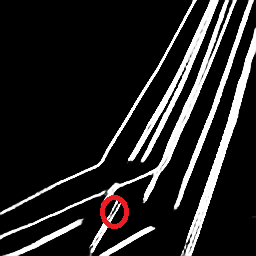}&\includegraphics[width=.14\linewidth]{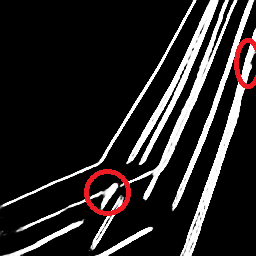}
&\includegraphics[width=.14\linewidth]{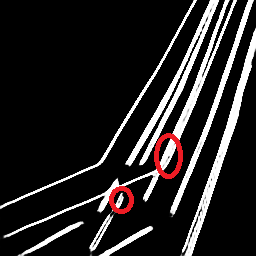}&\includegraphics[width=.14\linewidth]{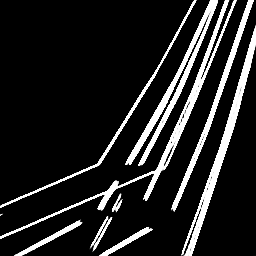}
&\includegraphics[width=.14\linewidth]{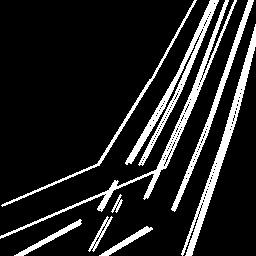}

							\\	
 \includegraphics[width=.14\linewidth]{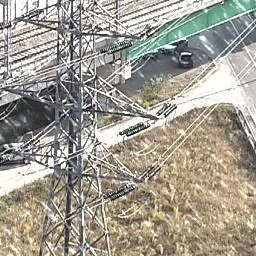}	&\includegraphics[width=.14\linewidth]{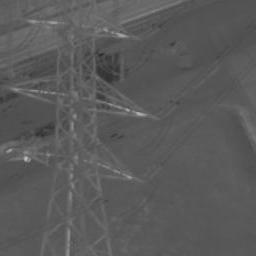}		
 &\includegraphics[width=.14\linewidth]{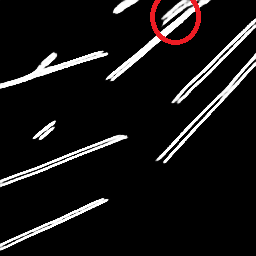}&\includegraphics[width=.14\linewidth]{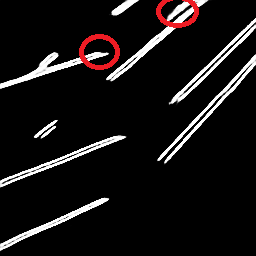}
&\includegraphics[width=.14\linewidth]{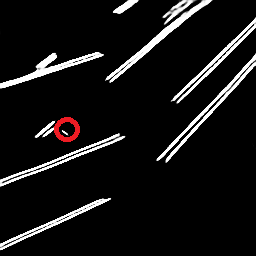}&\includegraphics[width=.14\linewidth]{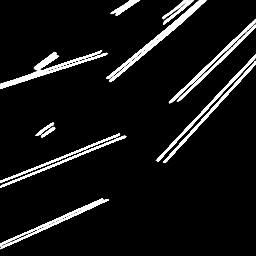}
&\includegraphics[width=.14\linewidth]{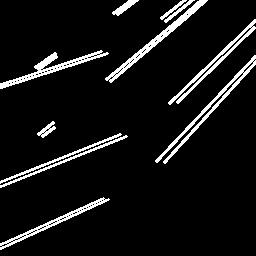}

							\\
		(a) RGB & (b) IR & (c) U-Net & (d) wMMEB & (e) wFAB&(e) HMMEN & (f) GTs\\
		
	\end{tabular}
	
	\caption{Visual detection examples of TL images using different configured models.}
	\label{fig:abla}
	\vspace{-4mm}
\end{figure*}

\begin{table}[h] \small
	\caption{Quantitative results of statistical tests based on Intersection over Union (IoU). The highest scores are highlighted in bold.}
    \vspace{5mm}
	\label{Tab:stest}
	\centering
	\begin{tabular}{c|c|c|c}
		\hline
	Method& Variance & Std. Dev.& Time (s)\\
            \hline
          U-Net &0.01891& 0.1375&0.0599
		\\
            \hline
           SegNet&0.02867&0.1693&\textbf{0.0481}
            		\\
            \hline
            MFNet&0.02867&0.1693&0.0610
            	\\
            \hline
            LHRNet&0.03378&0.1838&0.1008
            	\\
            \hline
           Bisenet&0.01237&0.1112&0.0501
            		\\
            \hline
            BGRNet&0.01731&0.1316&0.1760
            		\\
            \hline
		HMMEN& \textbf{0.00969}&\textbf{0.9838}&0.0660		\\
		\hline
	\end{tabular}
	\vspace{-2mm}
\end{table}

\begin{table}[h] \small
	\caption{Quantitative results of the one-sided t-test. }
    \vspace{5mm}
	\label{Tab:ttest}
	\centering
	\begin{tabular}{c|c|c|c|c|c}
		\hline
	 & U-Net & SegNet&MFNet&BGRNet&HMMEN\\
            \hline
          U-Net &0& 1&1&-1&-1
		\\
            \hline
           SegNet&-1&0&-1&-1&-1
            		\\
            \hline
            MFNet&-1&1&0&-1&-1
            	

            		\\
            \hline
            BGRNet&1&1&1&0&-1
            		\\
            \hline
		HMMEN& 1&1&1&1&0		\\
		\hline
	\end{tabular}
	\vspace{-2mm}
\end{table}

We conducted statistical testing based on Intersection over Union (IoU) to assess the robustness and reliability of our proposed method. Variance and standard deviation were utilized as metrics to evaluate the consistency and stability of performance across different methods. As shown in Table~\ref{Tab:stest}, our method achieved the lowest variance and standard deviation, indicating superior robustness compared to other approaches.

In addition, a one-sided t-test was performed at a 95\% confidence level to provide further statistical validation. The results are presented in Table~\ref{Tab:ttest}, where a value of '1' indicates that the method in the corresponding row outperformed the compared methods, and '-1' indicates underperformance. The results demonstrate that our method consistently outperformed other detection methods across all evaluations.

We also measured the runtime of each method on the same hardware to evaluate computational efficiency. The results, included in Table~\ref{Tab:stest}, indicate that our proposed method achieves comparable runtime performance to other methods while being significantly faster than BGRNet \cite{zhou2023transmission}. This balance of efficiency and robustness further highlights the practical advantages of our approach for transmission line detection in real-world applications.

\begin{table}[h] \small
	\caption{Quantitative detection results of state-of-the-art methods and the proposed method on RGB and IR transmission line (TL) images. The highest score for each metric is highlighted in bold. Results are presented as percentages for compactness.}
    \vspace{5mm}
	\label{Tab:fusion}
	\centering
	\begin{tabular}{c|c|c|c|c|c|c}
		\hline
	Method&  Day & Fog& Snow&Night& Old& Mean \\
            \hline
          U-Net &57.99 &55.91& 57.30&53.65&59.51&56.87
		\\
            \hline
           FuseNet&60.77 &59.32&  58.57&55.62&60.77&58.37
            		\\
            \hline
           MMTM&  59.61& 58.08& 59.20&57.39&60.95&59.04
            		\\
            \hline
            AMIFFM&60.89&59.00&60.59&58.00&61.71&60.03
            		\\
            \hline
		wMMEB& \textbf{66.82}& \textbf{63.55} & \textbf{65.01}&\textbf{61.62}&\textbf{68.58}&\textbf{65.11}		\\
		\hline
	\end{tabular}
	\vspace{-2mm}
\end{table}

Finally, we demonstrate the superior performance of our feature enhancement approach by comparing it with other state-of-the-art fusion strategies. The focus of this paper is on detecting transmission lines (TLs) from RGB and IR images, which heavily depends on effective fusion methods. To validate the generalization and robustness of the proposed feature enhancement module, we compared it with other state-of-the-art methods~\cite{joze2020mmtm,hazirbas2017fusenet,Choi2022attention}.
For a fair comparison, all experiments were conducted based on the U-Net architecture. We integrated the proposed feature enhancement module and other state-of-the-art fusion modules into U-Net, ensuring identical experimental settings for training. Following the methodology of AMIFFM~\cite{Choi2022attention}, we evaluated detection performance under diverse weather conditions, including haze, nighttime, daytime, and snow.

The results, as presented in Table~\ref{Tab:fusion}, clearly show that the proposed feature enhancement module achieves the highest detection performance, significantly outperforming other methods across all tested conditions. This substantial improvement highlights the effectiveness of the proposed module in leveraging complementary information from RGB and IR images, as well as its ability to generalize and maintain robustness under varying environmental challenges.

\subsection{Ablation Study}


\begin{table}[h] \small
	\caption{Quantitative detection results of state-of-the-art methods and proposed method on RGB and IR TL images.}
    \vspace{5mm}
	\label{Tab:aba}
	\centering
	\begin{tabular}{c|c|c|c|c|c|c}
		\hline
	Method& $S_e$ & Dice& $AUC$& AP& $P_r$ & IoU\\
            \hline
          U-Net &64.0 &65.5& 80.8&98.4&67.7&56.9
		\\
            \hline
           wMMEB&72.8 &71.9& 86.3&99.0&71.4&65.1
            		\\
            \hline
           wFAB&  70.5& 70.0 & 84.3&98.8&69.7&62.5
            		\\
            \hline
		HMMEN& \textbf{72.3}& \textbf{72.6} & \textbf{86.2} &\textbf{99.1}&\textbf{73.0}&\textbf{66.6}			\\
		\hline
	\end{tabular}
	
\end{table}

\begin{figure}[!htbp]
	\begin{center}
		\tabcolsep 1pt
		\begin{tabular}{@{}ccccc@{}}
			\includegraphics[width = 0.2\textwidth]{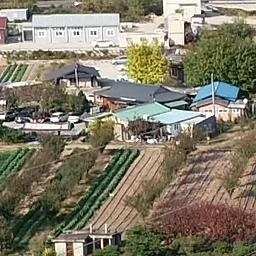} &
			\includegraphics[width = 0.2\textwidth]{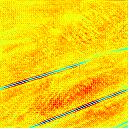}&
             \includegraphics[width = 0.2\textwidth]{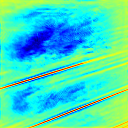} &
			\includegraphics[width = 0.2\textwidth]{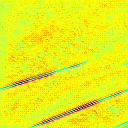}&
			\includegraphics[width = 0.2\textwidth]{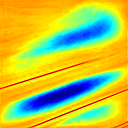}   \\
			
          \includegraphics[width = 0.2\textwidth]{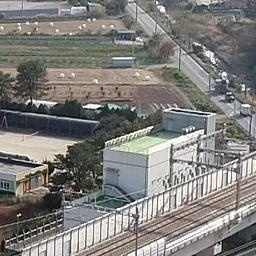} &
			\includegraphics[width = 0.2\textwidth]{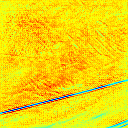}&
             \includegraphics[width = 0.2\textwidth]{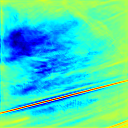} &
			\includegraphics[width = 0.2\textwidth]{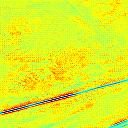}&
			\includegraphics[width = 0.2\textwidth]{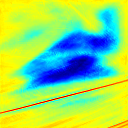}   \\

\includegraphics[width = 0.2\textwidth]{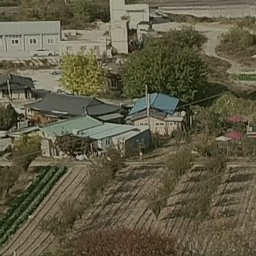} &
			\includegraphics[width = 0.2\textwidth]{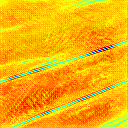}&
             \includegraphics[width = 0.2\textwidth]{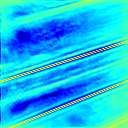} &
			\includegraphics[width = 0.2\textwidth]{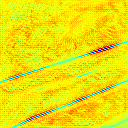}&
			\includegraphics[width = 0.2\textwidth]{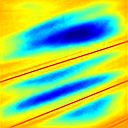}   \\

			Input &U-Net&wMMEB&wFAB&HMMEN\\
		\end{tabular}
	\end{center}
	\vspace{-3mm}
	\caption{Heatmap Visualization Demonstrating the Effectiveness of Proposed Modules. This figure highlights the heatmaps from the penultimate layer of different model configurations, showcasing the contributions of the proposed modules in enhancing feature representation and boundary clarity for transmission line detection.}

	\label{fig-heat}
\end{figure}

To evaluate the effectiveness of the proposed modules, we conducted experiments using various model configurations. First, the Mutual Multi-modal Enhanced Block (MMEB) was integrated into U-Net, resulting in the configuration named wMMEB. Next, the Feature Alignment Block (FAB) was incorporated into U-Net, creating the wFAB configuration. Finally, both MMEB and FAB were combined within U-Net, leading to the full model referred to as HMMEN. The detection performance of these configurations is summarized in Table~\ref{Tab:aba}.

The results demonstrate that while FAB contributes modest improvements in detection accuracy, MMEB significantly enhances the detection performance. Combining FAB with MMEB further refines the model's accuracy. Notably, all variants achieve comparable Average Precision (AP) scores; however, the Intersection over Union (IoU) scores reveal greater variation. As IoU places a stronger emphasis on correctly classifying positive pixels, it provides a more nuanced evaluation of detection quality. Consequently, we prioritize IoU for further analysis.

Detection results from the various configurations are visualized in Fig.~\ref{fig:abla}. The baseline U-Net model struggles with producing clear transmission line (TL) boundaries, leading to blurred edges. The wMMEB configuration mitigates this issue by improving the distinction between TLs and the background, although it still encounters difficulties in differentiating closely spaced lines. Meanwhile, the wFAB configuration occasionally introduces false positives. In contrast, the HMMEN configuration, incorporating both MMEB and FAB, not only delivers more precise TL detection but also effectively reduces false detections.

To further illustrate the impact of the proposed modules, we visualized the heatmaps of the penultimate layer from different model configurations, as depicted in Fig.~\ref{fig-heat}. The baseline U-Net produces heatmaps that are chaotic and lack a clear separation between transmission line (TL) features and the background. When the Mutual Multi-modal Enhanced Block (MMEB) is integrated, the heatmaps show significantly improved discriminability, effectively separating the TL features from the background clutter. Adding the Feature Alignment Block (FAB) further enhances the model's capability by producing sharper boundaries and better localization of TLs. These refinements are most evident in the heatmaps of the full HMMEN model, which demonstrate the synergistic effect of MMEB and FAB in recovering distinct and refined boundary features. Such improvements validate the effectiveness of the proposed modules in significantly enhancing the quality of TL detection.

\section{Conclusion}

In this paper, we introduce a novel and accurate transmission line detection (TLD) method built upon hierarchical multi-modal enhancement and feature alignment. The proposed approach leverages two key components, the Mutual Multi-modal Enhanced Block (MMEB) and the Feature Alignment Block (FAB), to address the inherent challenges of TLD. MMEB enhances the feature representation capability of multi-modal data by extracting and integrating the most useful features from RGB and infrared images. Simultaneously, FAB effectively mitigates the misalignment between high-level and low-level feature maps, ensuring precise feature integration and improved detection accuracy.
To validate the effectiveness of the proposed method, we conducted extensive experiments using the TLD dataset. The results demonstrate that our method consistently outperforms state-of-the-art TLD models in terms of detection accuracy and robustness, particularly under challenging weather and environmental conditions. Through a comprehensive ablation study, we quantified the individual contributions of MMEB and FAB, confirming their critical roles in improving detection performance. Visualizations of heatmaps further provided insights into the enhanced feature representations and refined boundaries facilitated by the proposed modules.
The proposed method not only achieves superior performance in detecting transmission lines but also establishes a robust framework for multi-modal data fusion and feature alignment, which can be extended to other applications in computer vision. Future work could focus on optimizing the computational efficiency of the proposed model and exploring its application in real-time scenarios, such as UAV-based power grid inspection. Overall, this work represents a significant step forward in leveraging hierarchical multi-modal enhancement and feature alignment for robust and accurate TLD.

{\flushleft\textbf{Declaration of competing interest}}

The authors declare that they have no known competing financial interests or personal relationships that could have appeared to influence the work reported in this paper.

{\flushleft\textbf{Data availability}}

Data will be made available on request.





\bibliographystyle{elsarticle-num}
\bibliography{egbib}

\end{document}